\documentclass[10pt,twocolumn,letterpaper]{article}
\usepackage{diagbox}
\usepackage[]{cvpr}      

\usepackage{graphicx}
\usepackage{amsmath}
\usepackage{amssymb}
\usepackage{booktabs}
\usepackage{amsfonts}
\usepackage{algorithm,algorithmic}
\usepackage{caption}
\usepackage{subcaption}
\usepackage[pagebackref,breaklinks,colorlinks]{hyperref}

\usepackage[capitalize]{cleveref}
\crefname{section}{Sec.}{Secs.}
\Crefname{section}{Section}{Sections}
\Crefname{table}{Table}{Tables}
\crefname{table}{Tab.}{Tabs.}

\def\cvprPaperID{7638} 
\def\confName{CVPR}
\def\confYear{2023}

\begin{document}

\title{Physical-World Optical Adversarial Attacks on 3D Face Recognition}

\author {
    Yanjie Li,\textsuperscript{\rm 1}
    Yiquan Li,\textsuperscript{\rm 1}
    Xuelong Dai,\textsuperscript{\rm 1}
    Songtao Guo,\textsuperscript{\rm 2}
    Bin Xiao\textsuperscript{\rm 1}\\
     \textsuperscript{\rm 1} Computing Department, The Hong Kong Polytechnic University \\
    \textsuperscript{\rm 2} College of Computer Science, Chongqing University\\  
    \{yanjie.li,yiquan.li,xuelong.dai\}@connect.polyu.hk, guosongtao@cqu.edu.cn,\\
    csbxiao@comp.polyu.edu.hk
}

\maketitle

\begin{abstract}
2D face recognition has been proven insecure for physical adversarial attacks. However, few studies have investigated the possibility of attacking real-world 3D face recognition systems. 3D-printed attacks recently proposed cannot generate adversarial points in the air. In this paper, we attack 3D face recognition systems through elaborate optical noises. We took structured light 3D scanners as our attack target. End-to-end attack algorithms are designed to generate adversarial illumination for 3D faces through the inherent or an additional projector to produce adversarial points at arbitrary positions. Nevertheless, face reflectance is a complex procedure because the skin is translucent. To involve this projection-and-capture procedure in optimization loops, we model it by Lambertian rendering model and use SfSNet to estimate the albedo. Moreover, to improve the resistance to distance and angle changes while maintaining the perturbation unnoticeable, a 3D transform invariant loss and two kinds of sensitivity maps are introduced. Experiments are conducted in both simulated and physical worlds. We successfully attacked point-cloud-based and depth-image-based 3D face recognition algorithms while needing fewer perturbations than previous state-of-the-art physical-world 3D adversarial attacks. 
\end{abstract}

\section{Introduction}
\label{sec:intro}
 In recent years, 3D face data are increasingly used for user authentication and other tasks to compensate for the weakness of 2D face recognition. Structured light imaging is one of the most popular 3D face data acquisition methods. It has high measurement precision and superiority in uniform textures \cite{feng2021calibration}, therefore is widely used in off-the-shelf devices such as Kinect v1 and iPhone X \cite{zhou20183d}. At the same time, many 3D face recognition algorithms are proposed \cite{gilani2018learning,kim2017deep,tan2019face}. Some of them have already been applied in the real world. For example, Apple's FaceID uses a structured light camera to collect 3D face data and then utilizes it for user identification.
   
\begin{figure}[t]
\centering  
{
\includegraphics[width=0.45\textwidth,height=0.18\textwidth]{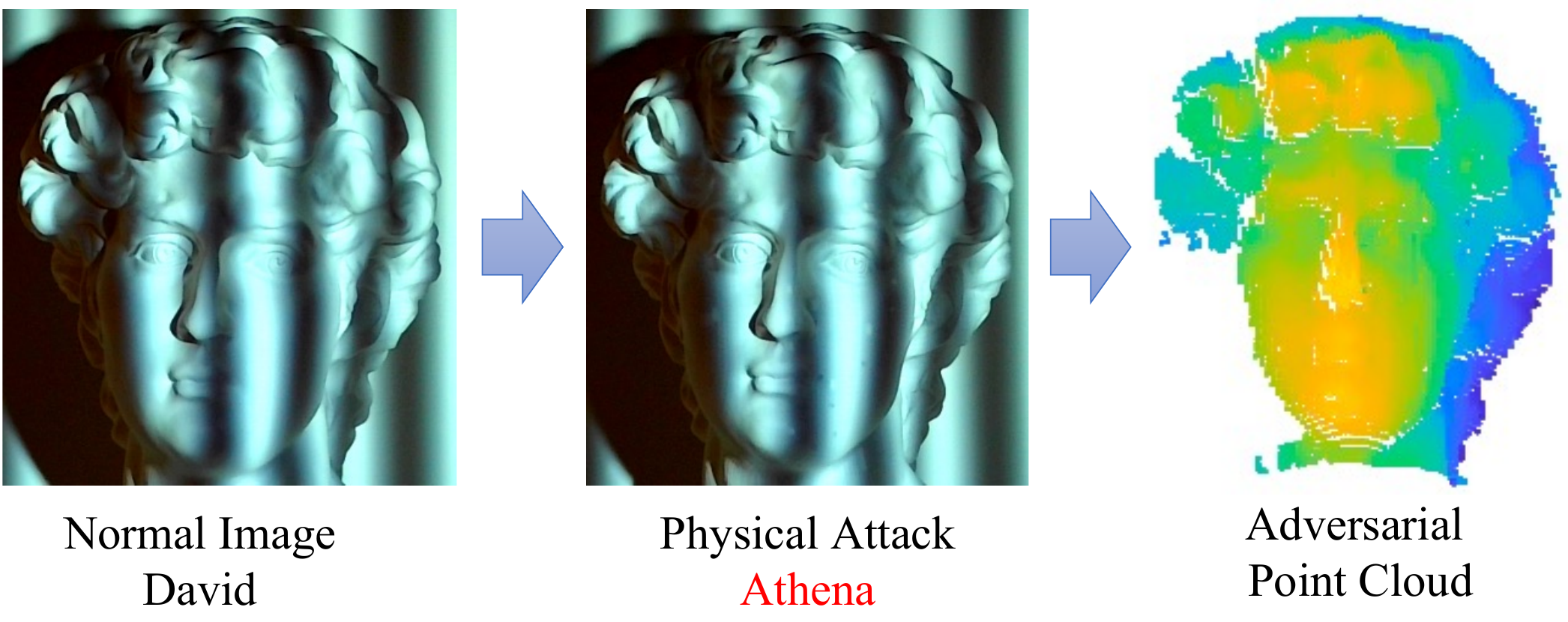}
} 
\caption{An demonstration of our attacks. We project additional noises on the fringe images to generate adversarial point clouds. Our attacks do not need the adversarial points strictly adjacent to the 3D surface and therefore need to modifier fewer points than state-of-the-art physical 3D adversarial attacks.}   
\label{overview}  
\end{figure}

Although real-world attacks for 2D face recognition have been thoroughly studied \cite{sharif2016accessorize,dong2019efficient,xiao2021improving}, studies on physical-realizable 3D face recognition attacks are still insufficient. The crux is that 3D face recognition is resistant to present 2D attack methods, such as adversarial patches. Tsai \textit{et al.} \cite{tsai2020robust} firstly proposed 3D printable adversarial examples for point cloud classification tasks, but because of the limitations of 3D printing technology, the perturbations must be strictly adjacent to the surface rather than at arbitrary positions in the 3D space. To solve these problems, inspired by optical adversarial attacks \cite{nguyen2020adversarial, nichols2018projecting, zhou2018invisible, worzyk2019physical,gnanasambandam2021optical}, this paper proposes novel physical adversarial attacks that use adversarial illumination to attack structured-light-based 3D face recognition system. The perturbations can be concealed in the normal fringe patterns or be superimposed on the original illumination by using the inherent or an additional projector. An overview of our attack is shown in Figure \ref{overview}. 

Nevertheless, because face reflection is more complex than opaque objects like the stop sign, and 3D reconstruction has great differences from 2D imaging, previous optical adversarial attacks cannot be applied directly for 3D face recognition. We propose to involve the 3D face relighting process in the attack pipeline through the Lambertian reflectance model. Then we propose a differential 3D reconstruction substitution. The perturbed light will result in point shiftings in the point cloud and finally, cause dodging or impersonation. Benefiting from the inherent advantages of optical adversarial perturbations, our methods can generate adversarial points at arbitrary 3D positions and therefore need fewer perturbations than previous SOTA 3D physical attacks with average attack success rates of 47\% of impersonation attacks and 95\% of dodging attacks. This paper's key contributions are summarized as follows.

    \begin{itemize}
        \item We realized physical adversarial attacks on the 3D face recognition through adversarial illuminations. The optical perturbations are generated end-to-end and are camouflaged by the normal fringe patterns of structured-light-based depth cameras.
        \item We involve the face relighting process in our attack pipeline through the Lambertian model. We also enhanced our physical attacks' invisibility and resistance to physical changes simultaneously with sensitivity maps and the 3D transformation-invariant loss.
        \item We attacked both point-cloud-based and depth-image-based 3D face recognition models and conduct ablation studies of different modules. Compared with SOTA 3D physical attacks, our method needs fewer noises without a cost in attack success rate.   
        \end{itemize}

    
\section{Related Work}
    \label{related_work}
    \paragraph{3D Adversarial Attack}
    3D adversarial attacks focus on generating adversarial examples for 3D deep learning models. Most of them aim for point cloud data. Xiang et al. \cite{xiang2019generating} first proposed an adversarial attack on the point cloud by point perturbation and point generation. Zheng et al. \cite{zheng2019pointcloud} and Wicker et al. \cite{Wicker_2019_CVPR} spoofed the deep learning model by dropping some critical points based on the saliency map. To improve the transferability of the attack, Hamdi et al. \cite{hamdi2020advpc} proposed advPC which adds an auto-encoder module into the adversarial point cloud generation process. Some other studies focused on improving the imperceptibility of the adversarial examples based on the geometric properties of the adversarial point cloud \cite{wen2022geometry, huang2022shape}. However, these attacks are hard to be realized in the physical world. 
        
    For physical realizable attacks, Cao et al. \cite{cao2019adversarial} successfully fooled the Lidar sensor on autonomous vehicles by adding fake front-near obstacles through a time-lapse module and a laser emitter. However, the sensors used for face recognition are significantly different from the autopilot. Some studies use 3D-printed objects for physical attacks. Tsai et al. \cite{tsai2020robust} proposed kNN loss to generate 3D printable adversarial examples. Tu et al. \cite{tu2020physically} proposed utilizing Laplacian loss to improve the mesh's smoothness and 3D printability. However, they did not apply their attacks to the 3D face recognition application, which is a far more complex scenario. Moreover, 3D printing technology can
    only generate perturbations adjacent to the 3D surface rather than at arbitrary positions in the 3D space. To the best of our knowledge, we are the first to realize physical adversarial attacks for 3D face recognition that can generate point-wise perturbations.     
    \paragraph{Optical Adversarial Attack}
    The optical adversarial attacks change the illumination of the target objects to spoof the classifiers. Compared with printing-based attacks, they have better camouflage. But previous works have only considered 2D classifiers. For example, Nicoles et al. \cite{nichols2018projecting} generated adversarial illuminations through iteratively capturing and optimizing. Zhou et al. \cite{zhou2018invisible} utilized infrared LEDs to attack the face recognition system. Worzyk et al. \cite{worzyk2019physical} projected perturbations onto the road stop signs. Nguyen et al. \cite{nguyen2020adversarial} applied the optical adversarial attack to the 2D face recognition and used expectation over transformation to improve the physical robustness. Gnanasambandam et al. \cite{gnanasambandam2021optical} improved attack success rate by considering spectral nonlinear. However, these attacks cannot be directly applied to 3D scenarios because of the huge difference in 2D and 3D imaging principles. Moreover, these attacks are too obvious to escape from human eyes. To solve these problems, this work proposes novel 3D optical adversarial attacks for face recognition scenario and consider the face reflection process for end-to-end attacks.

    \section{Methodology}
    \label{Section: Methodology}
    In this section, we first briefly introduce our attack targets, the structured light profilometry, and then introduces our attack methods. We propose phase shifting attack and phase superposition attack for multi-step and single-step technologies respectively. An overview of our end-to-end attacks is shown in Figure \ref{Fig-attack2_procedure}.

\subsection{Structured light profilometry}
    \label{subsection: Modeling_the_structure_light}      
    \paragraph{Multi-step algorithm} 
        \label{background}
          Multi-step phase shifting (MSPS) is one of the most important surface profilometry algorithms. We reference Piccirilli's work \cite{Piccirilli2016} to build the MSPS-based 3D face data acquisition system. We first calibrate the projector and camera to get their projection matrices $A_p$ and $A_c$. Then a group of phase-shift images are projected in sequential order to encode every unitary position of a surface. Next, the wrapped phase map $\phi_w$ is computed through captured images \(I_0^c,...,I_{N-1}^c\) by Equation \ref{eq_pha_wrapped},
            \begin{equation}
                \label{eq_pha_wrapped}
                \varphi_w(u_c, v_c)=\tan ^{-1} \frac{\sum_{n=0}^{N-1} I_{i}^c(u_c, v_c) \sin (2 \pi n / N)}{\sum_{n=0}^{N-1} I_{i}^c(u_c, v_c) \cos (2 \pi n / N)}.
            \end{equation}
        After getting the wrapped phase, it is unwrapped to the absolute phase, $\phi_a(u_c,v_c) = \phi_w(u_c,v_c) +2\pi K(u_c,v_c)$, where $K(u_c,v_c)$ is the fringe's order. We use cyclic complementary gray code \cite{wu2019high} to compute $K(u_c,v_c)$. Then pixels in $I_p$ and $I_c$ are matched according to the absolute phase. Finally, we use Eq.32 in Feng's paper \cite{feng2021calibration} to recover 3D coordinates, which, for the sake of simplicity, can be expressed as the following function,
            \begin{equation}
            \label{SLI_equation}
            [x,y,z] = h(A_c,A_p, u_c, v_c, u_p), 
            \end{equation}
        where $[x,y,z]$ is the 3D coordinate. The $u_c$, $v_c$ are the pixel's horizontal and vertical coordinates in $I_c$. $u_p$ is the matching pixel's horizontal coordinate in $I_p$.   

        \paragraph{Single-step algorithm} 
        Multi-step optical metrology faces the problems of error accumulation. Recently, benefiting from the rapid development of deep learning, some researchers proposed to project only one fringe image and then recover the phase map through deep neural networks \cite{zuo2022deep,qiao2020single,feng2019micro,feng2019fringe}. Feng et al. \cite{feng2019fringe} achieved SOTA results on single fringe pattern analysis. They rewrote Equation \ref{eq_pha_wrapped} as $\varphi_w(u_c, v_c)=\tan ^{-1}M(u_c, v_c)/D(u_c,v_c)$. They first used a CNN and a single modulated fringe image $I^c$ as input to estimate the background image $A$, and then used $A$ and $I^c$ as the second CNN's input to estimate $M$ and $D$ directly. We refer to this work as SLCNN in this paper.
        
    \subsection{Phase shifting attack}
\begin{figure}[t]
                \centering  
                {
                \includegraphics[width=0.32\textwidth,height=0.22\textwidth]{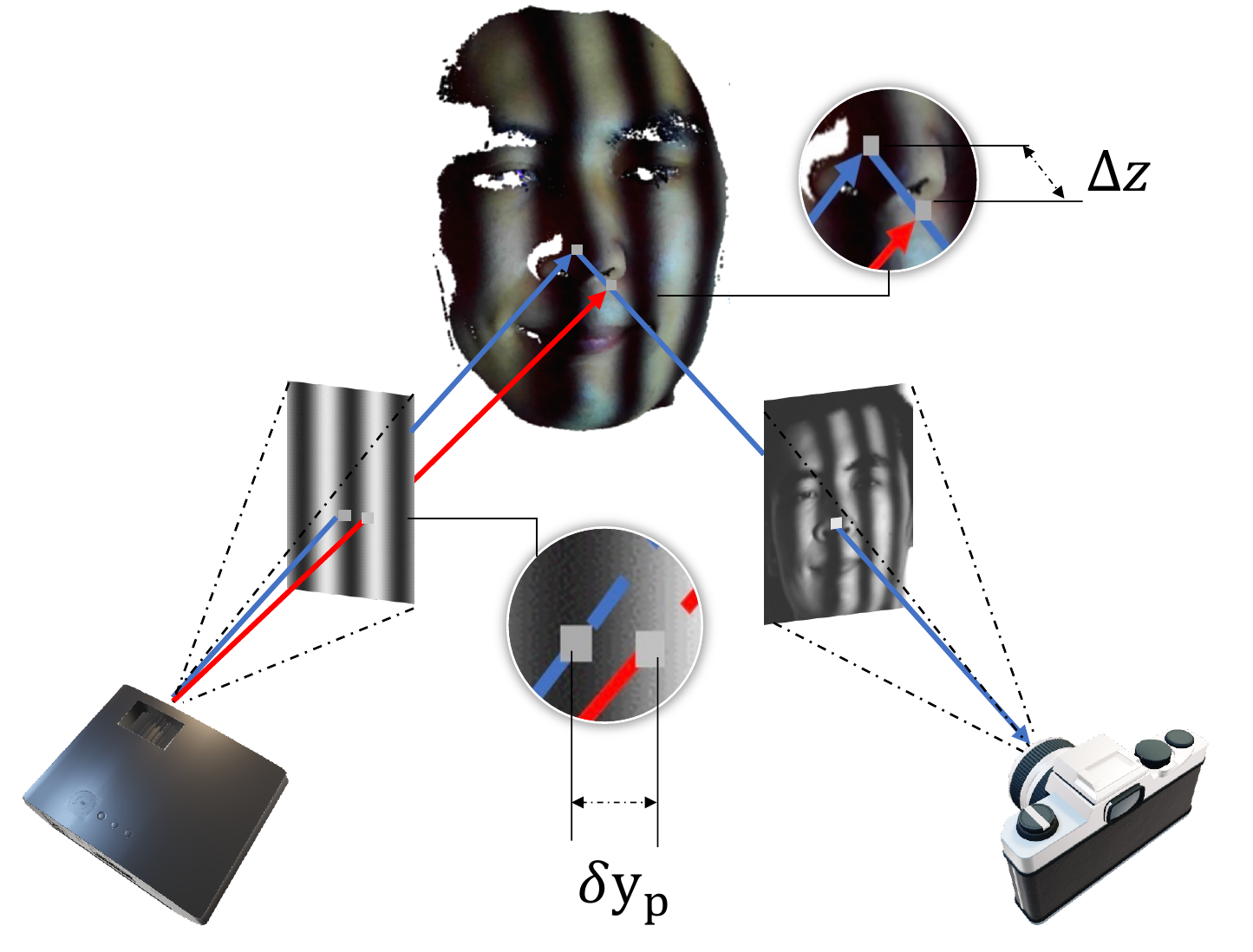}
                } 
                \caption{An overview of the phase-shifting attack. We modify the projected patterns to pollute the 3D data indirectly. The point displacements or depth changes are mapped to a flow field in the fringe image through a differential 3D reconstruction.}   
                \label{SLsystem}  
                \end{figure}

        For multi-step structured light imaging, we propose a novel attack named Phase Shifting attack, as shown in Figure \ref{SLsystem}. The basic idea is to involve the multi-step phase-shifting algorithm into the C\&W attack \cite{carlini2017towards}'s optimization process and hide the perturbation into the original patterns. However, because the pixel coordinates are discrete values, the back-propagation of Equation \ref{SLI_equation} is prevented. Therefore, we first substitute the 3D reconstruction algorithm with a differential one to solve this problem. 
    
            \paragraph{Differential 3D reconstruction algorithm}
            Inspired by the natural 3D renderer \cite{kato2018neural} that involves a differential renderer for 3D mesh reconstruction, we include the 3D reconstruction in the end-to-end attacks. The difference is that Kato's work needs to get $\frac{\partial I(u_c)}{\partial u_c}$, while our problem is to compute $\frac{\partial z(u_p,u_c)}{\partial u_p}$, which can be seen as an inverse rendering process. Specifically, because the pixel coordinate $u_p$ is not differential, we optimize the absolute phase map $\phi_a$ in the attack iterations instead of optimizing $u_p$ directly. After getting the adversarial phase map, we get $u_p$ though $u_p = round(\frac{w\phi_a}{2\pi n_s})$, where $n_s$ is the fringes' number, $w$ is the width of projected images.
            
            Moreover, simultaneously changing the projector's and the camera's corresponding points' coordinates will influence the reconstruction accuracy. To solve this problem, we project the 3D adversarial point displacements onto the normal vector of the camera imaging plane. Suppose $\boldsymbol{A_c}=K[R\; T] \in \mathbb{R}^{3 \times 4}$ is the camera's perspective projection matrix, then the projection direction can be expressed as $R^{-1}K^{-1}\boldsymbol{e}$, where $\boldsymbol{e}=(0,0,1)^T$. This projection will make the 3D pixel shifting only change the depth in the camera view and leave the corresponding pixel coordinates unchanged when the distance is in a certain range (we put the proof in the appendix). This is also equal to rotating the point cloud before the attack and limiting the perturbation in the depth direction.

    
            \paragraph{Sensitivity map and 3D transform invariant loss}   
             Noticing that humans are more sensitive to perturbation in the central and flat areas of the face, we propose two sensitivity maps to punish perturbations in areas of high sensitivity. The first is defined as $Sen1(u_c,v_c) = e^{-\frac{\|(u_c,v_c)-(u_0,v_0)\|_2}{w}}$, where $(u_0,v_0)$ is the face center. For the second map, we noticed the point cloud tends to be uniformly distributed on the surface after the farthest point sampling. When the point cloud is mapped to 2D images, the non-zero pixels are denser in areas with high depth changes. Therefore, the second sensitivity map is defined as $Sen2(u_c,v_c)= \frac{1}{\| \phi_a(u,v)\|_0}|_{(u,v)\in S(u_c,v_c)}$, where $S(u_c,v_c)$ is a small adjacent area centered by $(u_c,v_c)$ in the absolute phase map. The total sensitivity map is $Sen = Sen1 + Sen2$.

            When attacking a physical system, the distance and head may move unexpectedly. Expectation over transform loss has been used to solve this problem in 2D senario\cite{nguyen2020adversarial}. We extend it to the 3D space and propose 3D transform invariant loss to make the adversarial point clouds can generate consistent results when the environment changes. The random spatial transformation function can be expressed as
            $\mathcal{T}(\mathcal{P}) = (\textbf{R}_{(\theta_x,\theta_y,\theta_z)}\mathcal{P}^T)^T + \textbf{M}_{(\eta_x, \eta_y)}
            $, where $\textbf{M}$ and $\textbf{R}$ are spatial translation and rotation matrix and $\eta_x, \eta_y$, $\theta_x, \theta_y, \theta_z$ are random displacements and rotation angles sampled from
            Gaussian distributions. Moreover, to simulate the real-world preprocessing process, we involve a resample and renormalization function in the optimization. The final loss function is defined as
                \begin{equation}
                \label{Eq.3DTransform}
                l_{adv}(\phi_a') = \mathbb{E}_\mathcal{T}l_{logits}\left(\mathcal{M}\left(\mathcal{N}\left(\mathcal{T}(h(\phi_a'))\right)\right)\right)
                \end{equation}
            , where $\mathcal{T}$ is random 3D transformation. $\mathcal{N}(\cdot)$ is a resample and renormalization function. $\mathcal{M}$ is the classification model. $l_{logits}$ is a logits loss function \cite{carlini2017towards}.

     \paragraph{The phase shifting attack algorithm}    
           We use the $l_1$ distance as our distance loss to improve the sparsity of perturbation, which is defined as 
           $\|\phi_a'-\phi_a\|_1 = \sum_{u,v}|\phi_a'(u,v)-\phi_a(u,v)|$. Previous work has shown that $l_1$ loss function can generate sparser perturbation than $l_2$ loss \cite{modas2019sparsefool}. The final loss function is defined as
            \begin{equation}
            \label{Eq.totalloss}
            l_{total} = l_{adv} + \lambda \cdot Sen \odot \|\phi_a'-\phi_a\|_{1}     
            \end{equation}      
            , where $\odot$ is Hadamard products. We optimize this loss function by stochastic gradient descent with very small initial noises ($10^{-5}$) to skip the non-differential area. Last but not least, at the end of each iteration, we clip $\phi_a$ to $[max(0,\phi_a-1/n_s), min(1, \phi_a+1/n_s)]$. This is because when surfaces have large jumps, the multi-step phase-shift algorithm may unwrap the corresponding pixel into a false cycle. After getting the adversarial phase map, we shift the phase $I_p(u_p,v_p)$ in projected fringe patterns to $I_p(u_p',v_p)$ to get adversarial illuminations. The final attack algorithm is shown in Algorithm \ref{algorithm1}.

                \begin{algorithm}[htb]
                 \caption{Phase Shifting Attack Algorithm}
                 \begin{algorithmic}[1]
                 \renewcommand{\algorithmicrequire}{\textbf{Input:}}
                 \renewcommand{\algorithmicensure}{\textbf{Output:}}
                 \REQUIRE $I_c= \{I^c_1,...,I^c_N\}, I_p, A_c = K[R\;T]$, target label $t$
                 \ENSURE  $I_p^{adv}=\{I^p_1,...,I^p_N\}   $
                  \STATE $\phi_a \gets f(I_c)$
                  \FOR {$ i = 0$ to $N$}
                  \STATE Reconstruct the point cloud $\mathcal{P} \gets h(\phi_a)$
                  \STATE $\mathcal{P'} \gets  K \cdot R \cdot \mathcal{P}$
                  \STATE Compute the gradient $\nabla_{\mathcal{P}'}l_{adv}(\mathcal{P}', t)$
                  \STATE $\nabla_{\mathcal{P}'}l_{adv}(\mathcal{P}')[:,0:2] \gets 0$      
                  \STATE $\nabla_{\phi_a}l_{adv}(\phi_a) \gets \nabla_{\mathcal{P}'}l_{adv}(\mathcal{P}') \cdot \frac{\partial \mathcal{P}'}{\partial \phi_a}$
                  \STATE Compute the total gradient $\triangle = \nabla_{\phi_a}l_{total}(\phi_a)$  
                  \STATE $\phi_a \gets clip(\phi_a + \alpha \cdot \frac{\triangle}{\| \triangle \|_2} )$
                  \ENDFOR
                  \STATE $u_p' \gets round(\frac{w\phi_a}{2\pi n_s})$ 
                  \STATE $I_p^{adv}(u_p) \gets I_p(u_p') $    
                  \RETURN $I_p^{adv}=\{I^p_1,...,I^p_N\} $
                 \end{algorithmic} 
                 \label{algorithm1}
                 \end{algorithm}

        \begin{figure*}[t]
                \centering  \includegraphics[width=0.9\textwidth,height=0.3\textwidth]{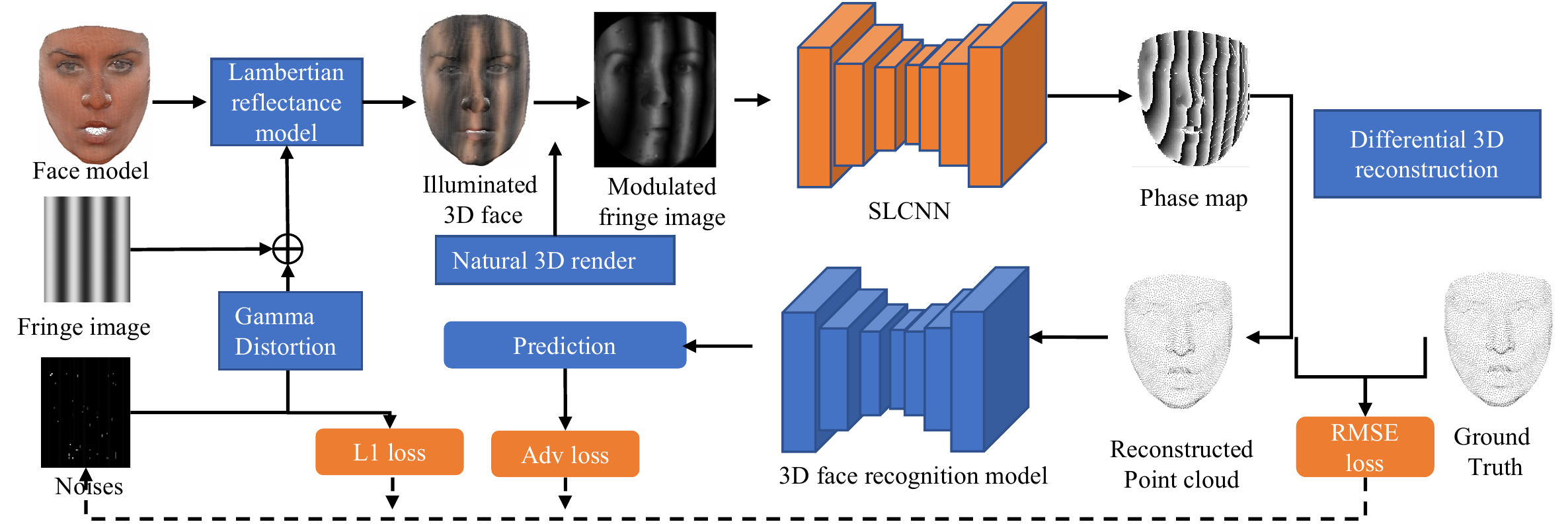}
                \caption{The end-to-end attack procedure of phase superposition attack. The fringe images and illumination noises relight the human face through the Lambertian reflectance model. Then a natural 3D renderer \cite{kato2018neural} renders the 3D model as modulated fringe images. 3D point clouds are then reconstructed by SLCNN and are classified through neural networks. Losses include adversarial loss, $L_1$ loss and RMSE loss that assesses the 3D face reconstruction accuracy.}
                \label{Fig-attack2_procedure}
        \end{figure*}   
        
        \subsection{Phase superposition attack}
        In the real world, the adversary usually cannot directly modify the inherent projector of a structured light system. Therefore, we propose another end-to-end attack called Phase Superposition Attack. As shown in Figure \ref{Fig-attack2}, the adversary uses an additional projector to project adversarial noises on the faces, resulting in dodging or impersonation attacks. We suppose the victim model is SLCNN \cite{feng2019fringe} or any other single-fringe-analysis neural network. 
    
        \paragraph{The Lambertian rendering model}
        Our end-to-end attack procedure is shown in Figure \ref{Fig-attack2_procedure}. In order to generate adversarial noises directly, we involve the projection-and-capture process in our optimization loops. Unfortunately, face reflection is a very complex process because of its translucent quality \cite{smith2010estimating}, which hasn't been considered in previous optical adversarial attacks. We use the linear Lambertian rendering model \cite{basri2003lambertian} to simulate this procedure, which has been used for illumination-invariant face recognition \cite{zhou2007appearance}. The linear Lambertian model is
            \begin{equation}
                I(\boldsymbol{x})= a(\boldsymbol{x})\boldsymbol{n}(\boldsymbol{x})\cdot(\boldsymbol{s}_{p_1}(\boldsymbol {x})+\boldsymbol{s}_{p_2}(\boldsymbol {x}))    
            \end{equation}
        , where $\boldsymbol{x}$ is the 3D face, $I(\boldsymbol {x})$, $a(\boldsymbol{x})$, $\boldsymbol{n}(\boldsymbol{x})$, $\boldsymbol{s_p}(\boldsymbol {x})$ are face's intensity, reflectance, the surface normal and light vector from projector $p$ respectively. We estimate reflectance and surface normal parameters through SfSNet \cite{sengupta2018sfsnet}, which can decompose a face image as $a(\boldsymbol{x})$,$\boldsymbol{n}(\boldsymbol{x})$,$\boldsymbol{s}(\boldsymbol {x})$ under Lambertian model through a single forward propagation. 
       
        \paragraph{The phase superposition attack algorithm}
        To both fool 3D face recognition models and make the recovered 3D face look realistic, we add a 3D Root Mean Square Error (RMSE) to measure the distance from the adversarial point cloud to the ground truth, which has been used for measuring 3D reconstruction accuracy \cite{liu2018disentangling}. RMSE is defined as $RMSE=\frac{1}{N_T}\Sigma_{i=1}^{N_T}(\|\mathcal{P^*}-\mathcal{P}\|/n)$, where $\mathcal{P^*}$ and $\mathcal{P}$ are the adversarial and normal point cloud respectively and $N_T$ is the attack batch size. To calculate RMSE, we align the adversarial and real 3D faces with 68 3D landmarks and then cropped them into the same radiuses. We also consider the projector's gamma distortion in our attacks, which is modeled as a tanh function in this paper. The phase superposition attack algorithm is shown in \ref{algorithm2}.
                 \begin{algorithm}[htb]
                 \caption{Phase Superposition Attack Algorithm}
                 \begin{algorithmic}[1]
                 \renewcommand{\algorithmicrequire}{\textbf{Input:}}
                 \renewcommand{\algorithmicensure}{\textbf{Output:}}
                 \REQUIRE Test 3D face data $P$, phase reconstruction model $\mathcal{M}_1$, 3D classification model $\mathcal{M}_2$
                 \ENSURE  Adversarial illumination $\boldsymbol{x}$
                 \\ \textit{Initialization} : $\boldsymbol{x} \gets \boldsymbol{1} \times 10^{-5}$                //\textit{Skip the zero}
                  \FOR {$ i = 0$ to $N$}
                  \STATE Relight the face through the Lambertian model
                  \STATE Get modulated images $I_p'$ by natural 3D renderer.
                  \STATE Get phase map through $\mathcal{M}_1$
                  \STATE Get $\mathcal{P^*}$ through differential 3D reconstruction
                  \STATE $RMSE \gets \frac{1}{N_T}\Sigma_{i=1}^{N_T}(\|\mathcal{P^*}-\mathcal{P}\|/n)$
                  \STATE Get the prediction through $\mathcal{M}_2$                  
                  \STATE Compute the adversarial loss $l_{adv}$
                  \STATE $l_{total} \gets l_{adv} + \lambda_1\cdot RMSE + \lambda_2 \cdot Sen \odot \|x\|_1$
                  \STATE Compute gradient $\triangle = \nabla_{\boldsymbol{x}}l_{total}(\boldsymbol{x})$  
                  \STATE $\boldsymbol{x} \gets max\{0, min\{1, \boldsymbol{x} + \alpha \cdot sign(\delta)\}\}$
                  \ENDFOR   
                  \RETURN Adversarial illumination $\boldsymbol{x}$
                 \end{algorithmic} 
                 \label{algorithm2}
                 \end{algorithm}        

        \section{Implementation details}
        \paragraph{Datasets}
            We first train 3D face recognition models on 3D face datasets for attack evaluation. Three datasets acquired by structured light cameras are used: Bosphorus \cite{savran2008bosphorus} and Eurecom \cite{min2014kinectfacedb} and SIAT-3DFE \cite{9028192}. The Bosphorus dataset is collected by Inspeck Mega Capturor II. It consists of 105 different faces with a rich set of expressions and occlusions. Eurecom consists of 52 faces and was acquired by Kinect. SIAT-3DFE consists of over 400 subjects and has original fringe images. We also collect ten people's faces using our own structured light camera and add them to the above datasets. We downsample these datasets to 4K points through farthest point sampling to train 3D classifiers.   
       \paragraph{3D face recognition models}     
            We evaluate our attacks on state-of-art 3D point cloud classification models, including Pointnet\cite{qi2017pointnet}, Pointnet++ \cite{qi2017pointnet++}, DGCNN\cite{wang2019dynamic} and CurveNet \cite{xiang2021walk}. We also evaluate attacks on the depth-image-based 3D face recognition model FR3DNet \cite{gilani2018learning}. We implemented dodging and impersonation attacks on these models. The dodging attack aims to make 3D classifiers classify faces into any classes except the ground truth. The impersonation attack aims to make them classify faces into randomly chosen labels.
    \paragraph{Hyperparameter settings}
           The search space of $\lambda$ in Equation \ref{Eq.totalloss} is set as [1e-5,1e5]. We use binary search to narrow down $\lambda$. We set the binary search step as 10, the iteration number as 100, and the minimum logit difference as 30 to guarantee the attack success rate (ASR). We terminate the optimization process after the binary search step is reached and return the adversarial examples with minimum distance loss. We use ASR and RMSE distance between the adversarial and original point clouds as evaluation metrics.  
               
            \begin{figure}[h]
                \centering  \includegraphics[width=0.4\textwidth,height=0.25\textwidth]{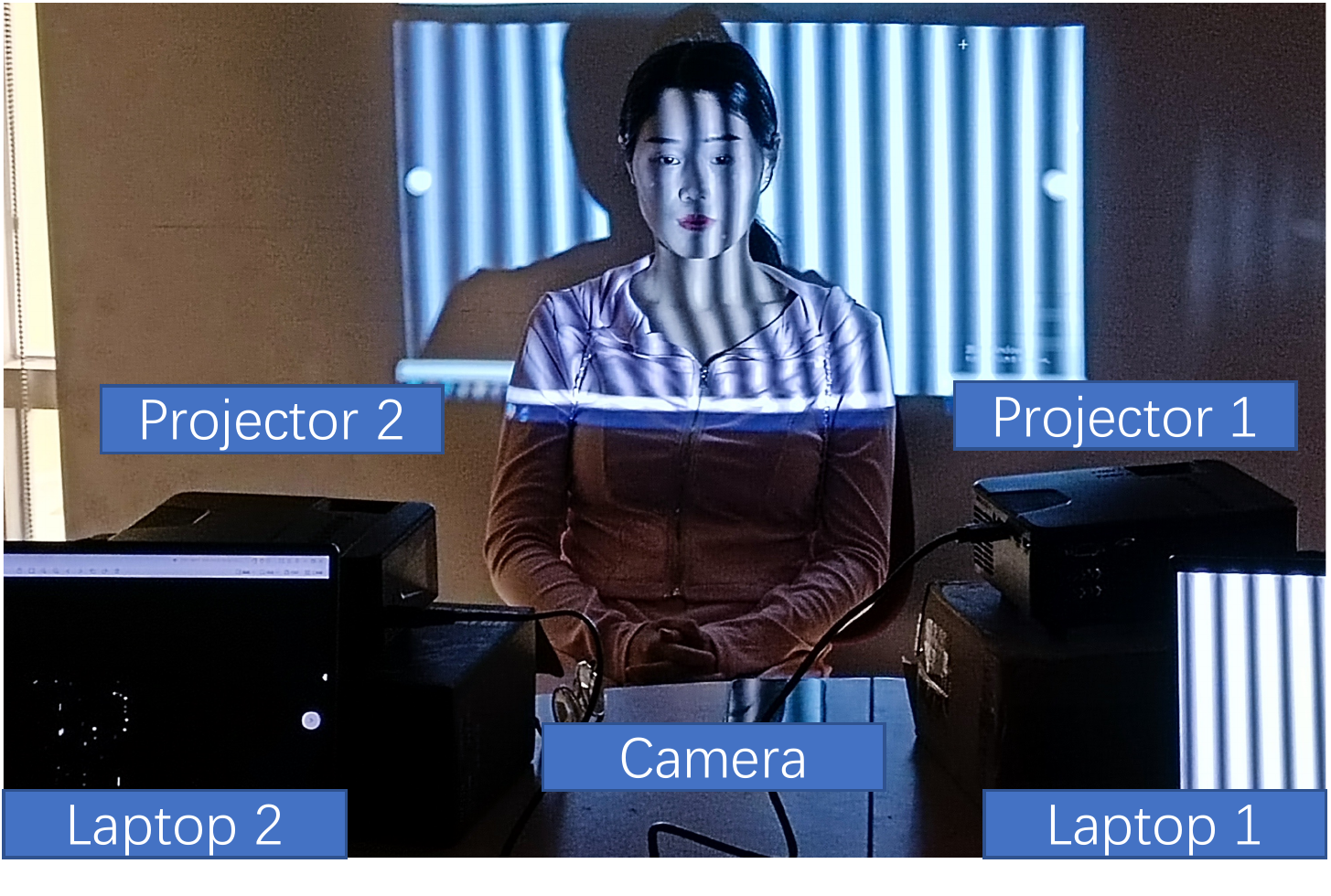}
                \caption{The physical settings of the phase superposition attack. The adversary uses an additional projector to add noises to the original fringe images for 3D reconstructions.}
                \label{Fig-attack2}
            \end{figure}    
                    
            \begin{figure*}[htbp]
                \centering  \includegraphics[width=0.96\textwidth,height=0.52\textwidth]{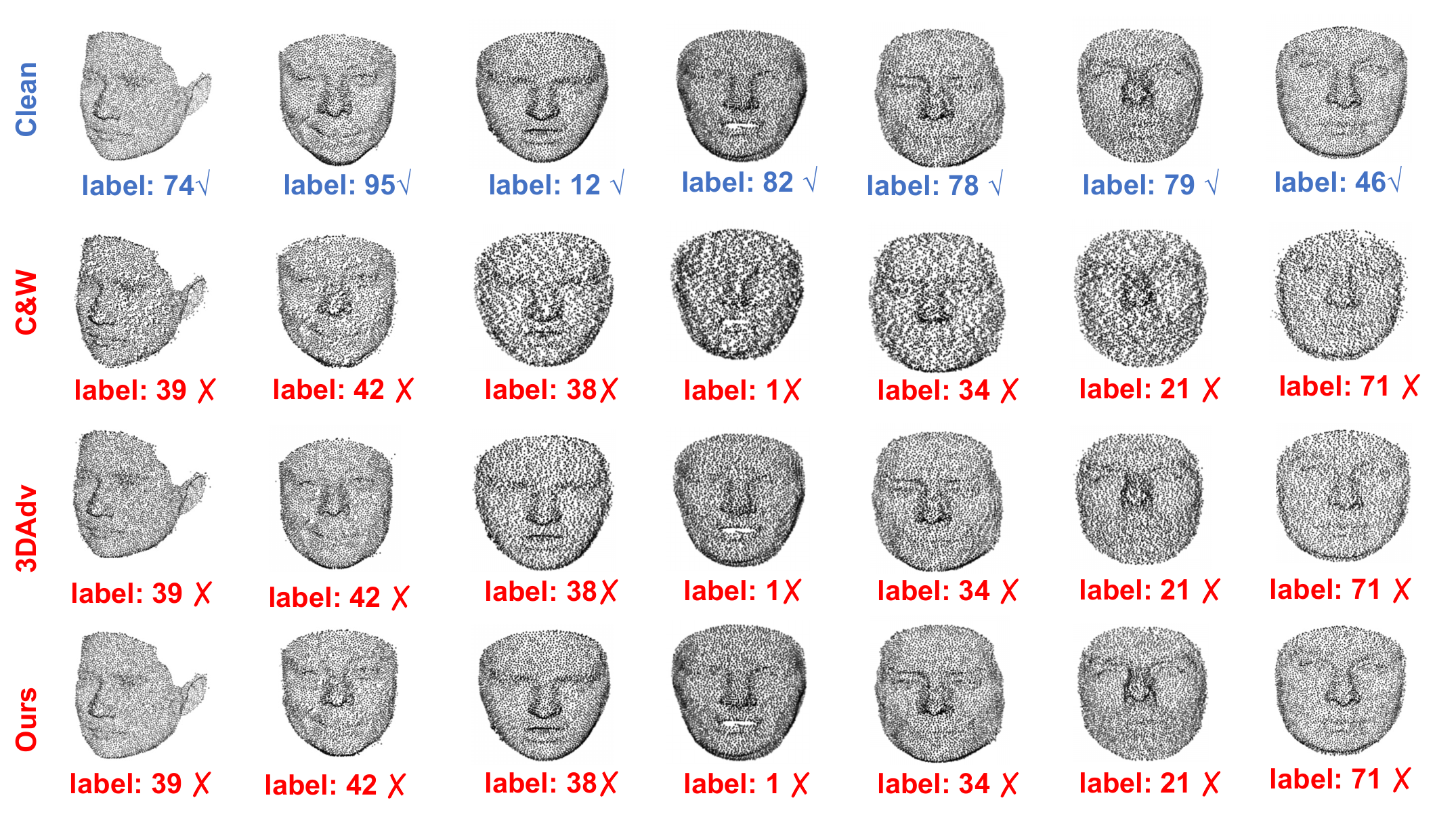}
                \caption{Demonstrations of clean and adversarial face point clouds generated by C\&W, 3DAdv and ours attacks. }
                \label{Fig:digital-attack}
                \end{figure*}                                     
            \paragraph{Physical settings}
               
            As shown in Figure \ref{Fig-attack2}, this paper built a real-world structured light system that refers to Piccirilli’s design \cite{Piccirilli2016} for physical attacks, which includes an industry camera and home projectors. We capture images at a distance of about 1.5 meters. The projector and the camera's resolutions are both $1600\times1200$. We linearize the color distortion before the images are projected, then detect and crop the face in the modulated images using the Viola-Jones algorithm and resize it to $128\times 128$ centered by the face. We reconstruct the 3D data using the 12-step phase-shift algorithm or deep-learning-based methods such as SLCNN.
                       
            The projected images may be distorted because of manufacturing errors in the projector-camera system. This nonlinear distortion will cause a drop in the attack success rate if not taken into account. The luminance non-linearity usually are formulated as $g(u)=u^{\gamma}$, where $u\in [0,1]$ is the input intensity of the projector and $g(u)$ is the output intensity of the projector \cite{951529}. However, our experiments found that the $tanh$ function can better fit our projectors' distortion, which is $g(u) = \frac{tanh(\gamma \cdot (2u-1))+1}{2}$, where $\gamma$ is a hyper-parameter to describe the projector's inherent property. We leave the details about computing $\gamma$ in the appendix.

            \section{Experimental evaluation}
            We implemented attacks in both the digital world and the physical world. We present our experiments on 1) targeted and untargeted attacks on various classification backbones, 2) physical attack results, 3) ablation study of different modules, and 4) attack transferability.

             \begin{table*}[htbp]
                  \centering
                  \small
            \begin{tabular}{ccccccccccc}
            \toprule
               & \multicolumn{2}{c}{PointNet} & \multicolumn{2}{c}{PointNet++(SSG)} & \multicolumn{2}{c}{Point++(MSG)} & \multicolumn{2}{c}{DGCNN} & \multicolumn{2}{c}{CurveNet} \\
            \midrule
            Metrics & ASR (\%) & RMSE & ASR (\%) & RMSE & ASR (\%) & RMSE & ASR (\%) & RMSE & ASR (\%) & RMSE \\
            \midrule
            C\&W \cite{carlini2017towards} & 0.98  & 0.35  & 0.92  & 0.45  & 0.88  & 0.36  & 0.94  & 0.26  & 0.88  & 0.43  \\
            3Dadv \cite{xiang2019generating} & 0.89 & 0.27  & 0.86 & 0.34  & 0.77  & 0.25  & 0.89  & 0.16  & 0.95  & 0.36  \\
            KNNadv \cite{tsai2020robust} & 0.86  & 0.15  & 0.85  & 0.22  & 0.79  & 0.23  & 0.85  & 0.08  & 0.86  & 0.17  \\
            GeoA3 \cite{wen2020geometry} & 0.75  & 0.18  & 0.91  & \textbf{0.14} & 0.85  & 0.13  & 0.84  & 0.09  & 0.97  & 0.16  \\
            \textbf{Ours} & \textbf{0.95}  & \textbf{0.13} & \textbf{0.93}  & 0.15  & \textbf{0.99} & \textbf{0.11} & \textbf{0.96} & \textbf{0.05} & \textbf{0.97} & \textbf{0.09} \\
            \bottomrule
            \end{tabular}%
            
             \caption{The untargeted attack performance. We evaluate the attack performance by the attack success rate (higher is better) and RMSE \cite{liu2018disentangling} (lower is better). RMSE can be used to measure the 3D reconstruction error. Compared with the state-of-art geometry-ware attack GeoA3, our attacks have fewer 3D reconstruction errors while maintaining a high attack success rate. The RMSE is multiplied by 10 for comparison.}  
                \label{table-untarget-attack}  
            \end{table*}

            \begin{table*}[htbp]
                  \centering
                  \small
                \begin{tabular}{ccccccccccc}
                \toprule
                   & \multicolumn{2}{c}{PointNet} & \multicolumn{2}{c}{PointNet++(SSG)} & \multicolumn{2}{c}{Point++(MSG)} & \multicolumn{2}{c}{DGCNN} & \multicolumn{2}{c}{CurveNet} \\
                \midrule
                Metrics & ASR (\%) & RMSE & ASR (\%) & RMSE & ASR (\%) & RMSE & ASR (\%) & RMSE & ASR (\%) & RMSE \\
                \midrule
                C\&W \cite{carlini2017towards} & 0.57 & 0.76 & 0.42 & 0.79 & 0.32 & 0.81 & 0.43 & 0.69 & 0.48 & 0.88 \\
                3Dadv \cite{xiang2019generating} & 0.58 & 0.65 & 0.39 & 0.66 & 0.35 & 0.78 & 0.37 & 0.45 & 0.53 & 0.65 \\
                KNNadv \cite{tsai2020robust} & 0.52 & 0.54 & \textbf{0.45} & 0.51 & \textbf{0.38} & 0.54 & 0.16 & 0.34 & 0.27 & 0.49 \\
                GeoA3 \cite{wen2020geometry} & 0.45 & 0.52 & 0.35 & 0.54 & 0.24 & 0.45 & 0.28 & 0.31 & 0.25 & 0.42 \\
                \textbf{Ours} & \textbf{0.62} & \textbf{0.33} & 0.37 & \textbf{0.32} & 0.26 & \textbf{0.44} & \textbf{0.55} & \textbf{0.25} & \textbf{0.57} & \textbf{0.39} \\
                \bottomrule
                \end{tabular}%
             \caption{The targeted attack performance on our method and the other four attacks. The RMSE is multiplied by 10 for comparison.} 
                \label{table-target-attack}  
            \end{table*}

    \subsection{Attack performance}
            
             \paragraph{Digital attack results} To evaluate the targeted and untargeted attack success rate, we first simulate attacks through the Lambertian rendering model and then attack different models that are trained on three datasets. For targeted attacks, the label is randomly chosen from three datasets. We compared our attacks with several state-of-the-art 3D attacks, including C\&W attack \cite{carlini2017towards} that uses $l_2$ distance, 3Dadv \cite{xiang2019generating} that uses Chamfer distance, KNNadv \cite{tsai2020robust} that use kNN distance and GeoA3 \cite{wen2020geometry} that use local curvature distance. Note that recently shape-invariant attack has been proposed \cite{huang2022shape}, but it is not physically realizable, therefore is not considered in this paper. 

             As revealed in Table \ref{table-untarget-attack} and Table \ref{table-target-attack}, we achieved an average ASR of 95\% on dodging attacks and 47\% on impersonation attacks. At the same time, our attacks have fewer 3D reconstruction errors than previous 3D attacks. This is because our attacks need fewer points to be modified and consider the physical constraint in the 3D reconstruction by projecting the perturbation in specific directions. For the targeted attack, we outperform previous attacks except for KNNadv. We think this is may because KNNadv can generate adjacent clusters, which are more robust to the downsampling process of Pointnet++. We also visualize the adversarial point clouds in Figure \ref{Fig:digital-attack}.   

                \begin{figure}[hptb]
                \centering  \includegraphics[width=0.43\textwidth]{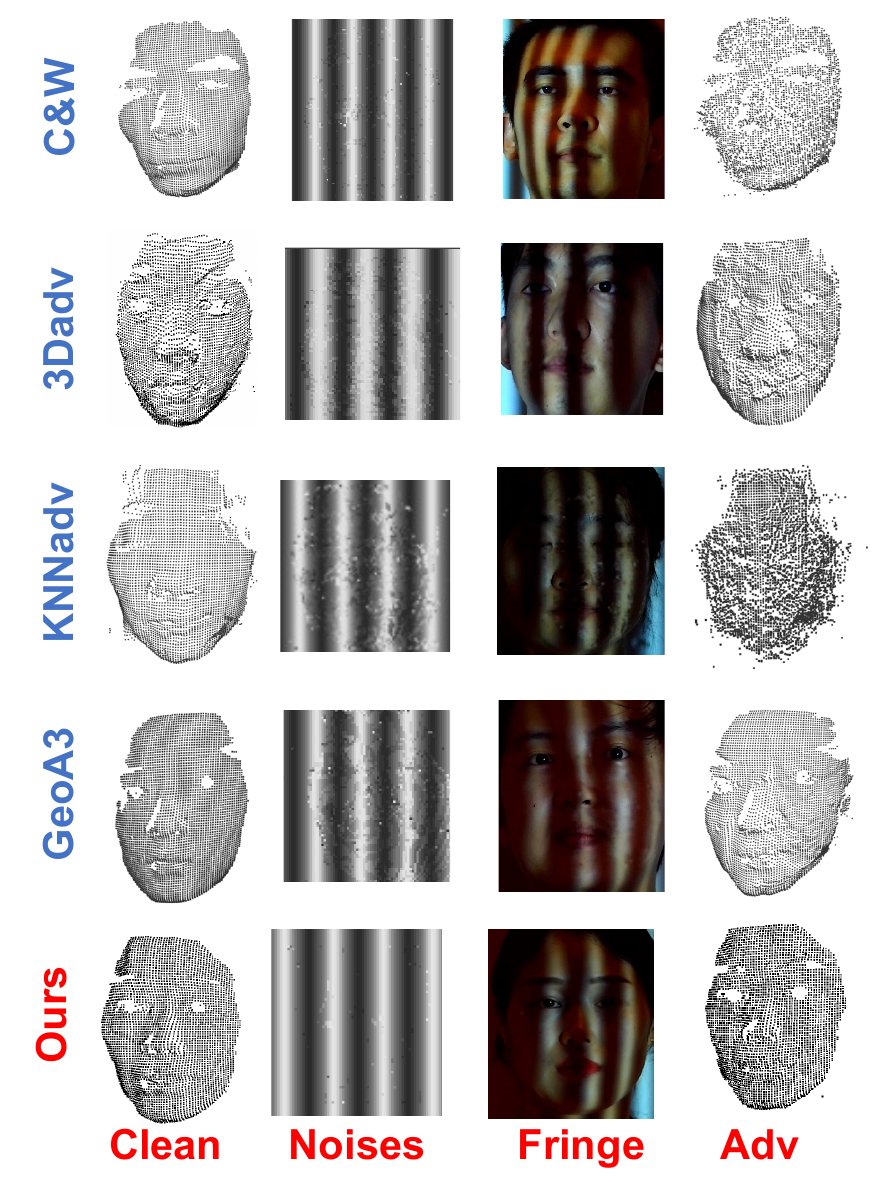}
                \caption{The physical phase shifting attack results. We apply previous attacks in the structured light system for comparison. \textbf{The first four rows}: benchmark attacks. \textbf{The last row}: Our attack.} 
                \label{Fig:pointcompare}
                \end{figure}
                
\begin{figure}[h]
            	\centering
            	\begin{subfigure}{0.3\linewidth}
            		\centering
            		\includegraphics[width=0.95\linewidth]{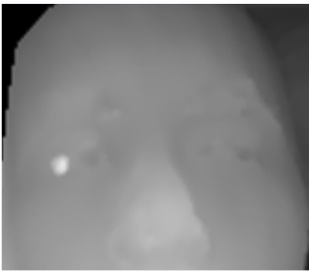}
            		\label{depth1}
            	\end{subfigure}
            	\centering
            	\begin{subfigure}{0.3\linewidth}
            		\centering
            		\includegraphics[width=0.95\linewidth]{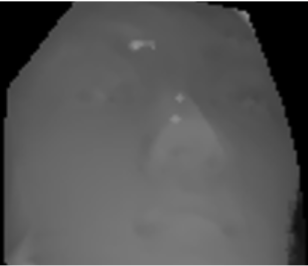}
            		\label{depth2}
            	\end{subfigure}

            	\caption{Adversarial examples for FR3DNet.}
            	\label{Fig-depth}
            \end{figure}

    \paragraph{Attack FR3DNet} 
              Because our attacks can generate points at arbitrary positions, they can also be applied to depth-image-based 3D face recognition. FR3DNet \cite{gilani2018learning} is a deep CNN model designed for 3D face recognition which uses depth information as input. The depth image can be easily recovered from the phase map so only minor modifications are required for our attacks. Figure \ref{Fig-depth} shows adversarial examples generated for FR3DNet.

    \paragraph{Physical attack results} 
              Figure \ref{Fig:pointcompare} shows the real-world attacks' results compared with other attacks. We show the phase superposition attack's results in the appendix. To compare with benchmark attacks, we use their loss functions to generate the 3D adversarial point clouds and then reverse them to the fringe patterns. We conduct targeted attacks on PointNet with randomly chosen labels. The results show that the reconstructed 3D point clouds by our attacks are more similar to real faces and few perturbations are needed for projection.
             
        \subsection{Ablation study}    
\begin{figure*}[htbp]
                   \centering
            	\begin{subfigure}{0.3\textwidth}
                    \includegraphics[width=1\textwidth]{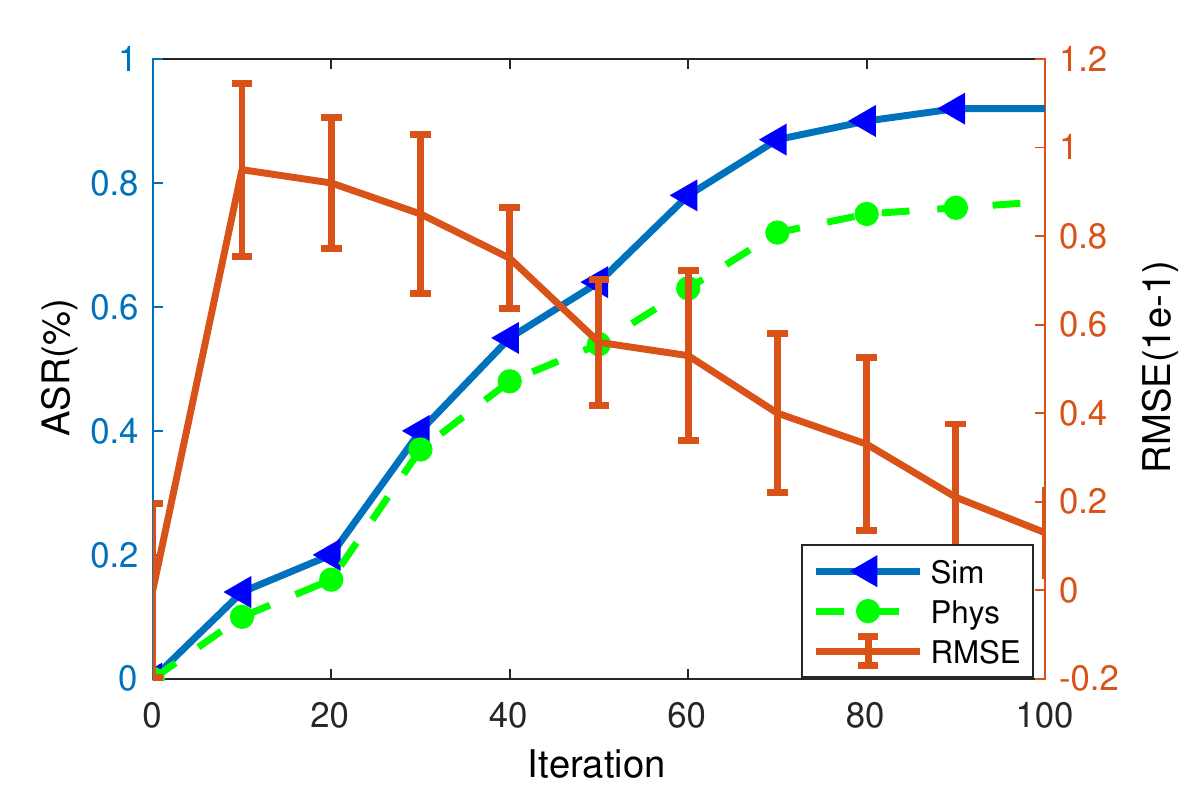}
            		\caption{Iteration}
            		\label{fig-iteration}
            	 \end{subfigure} %
              \centering
            	\begin{subfigure}{0.3\textwidth}
                    \includegraphics[width=1\textwidth]{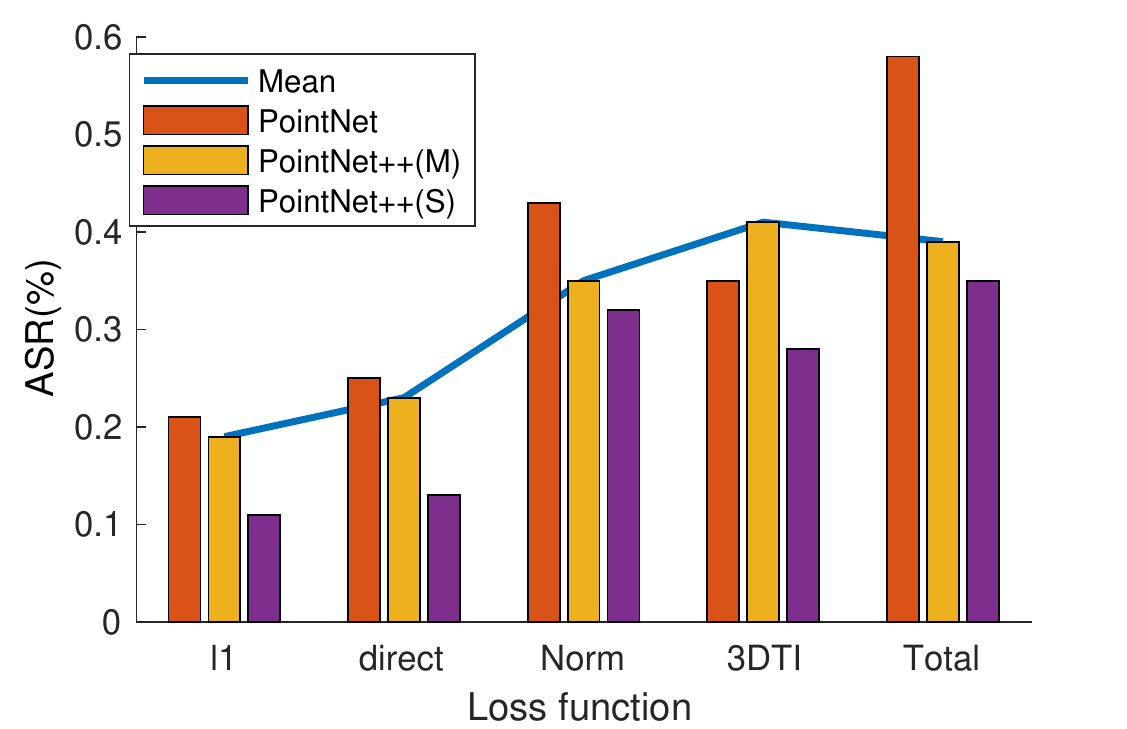}
            		\caption{ASR}
            		\label{fig-asr}
            	\end{subfigure} 
             \centering
            	\begin{subfigure}{0.3\textwidth}
                        \includegraphics[width=1\textwidth]{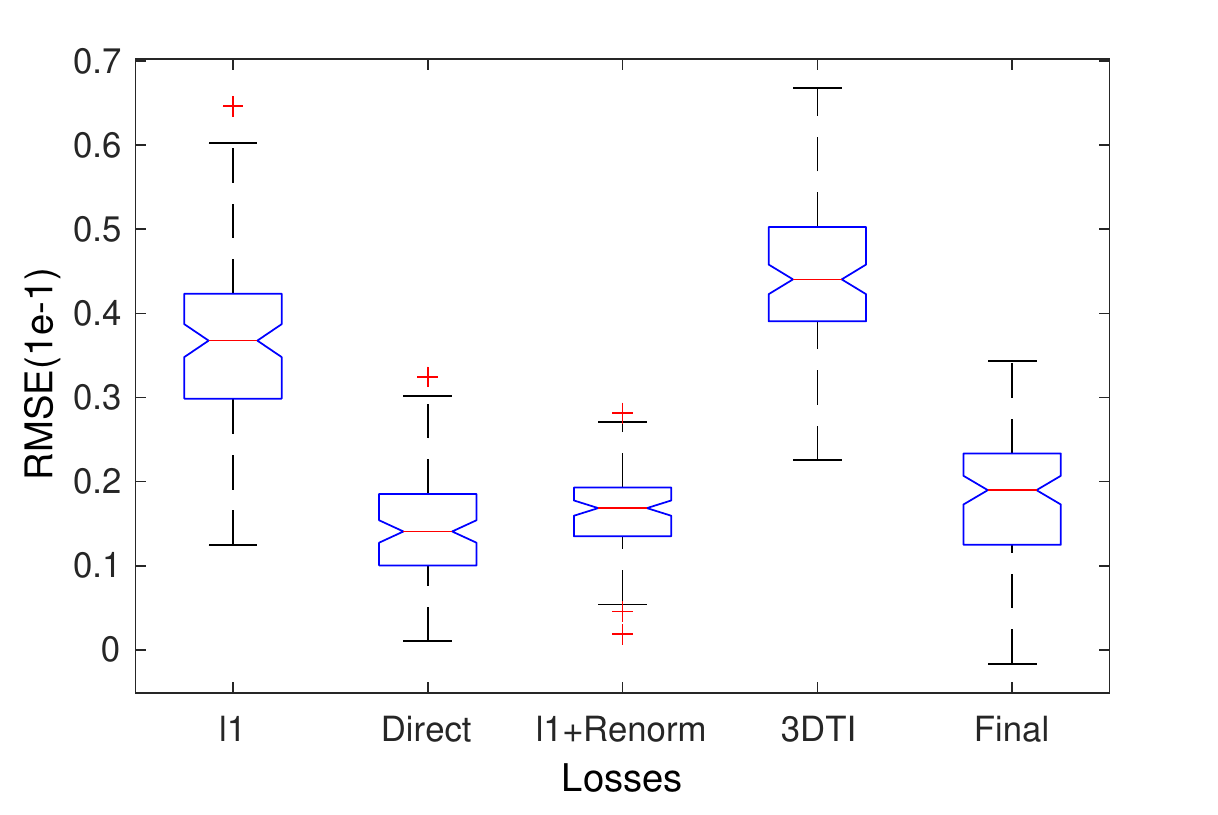}
            		\caption{RMSE}
            		\label{RMSE}
            	\end{subfigure}
                 \caption{The results of ablation study. $\Delta z$ means the direction constraint. $\mathcal{N}$ is the renormalization. $\mathcal{T}$ means random transformations.}
                 \label{Ablation-study}
            \end{figure*}

              \paragraph{Quantitative results of different modules} To evaluate the effects of different modules, we involve the random rotations and translations of the head and necessary preprocess steps (normalization and down-sampling) in the simulated and physical 3D reconstruction system, and evaluate the ASR on five different models (Figure \ref{fig-asr}). The naive $l_1$ attack without any techniques suffers a low ASR. With the effectiveness of direction constraint, renormalization, and 3D-TI, the targeted ASR increases by 34\% on Pointnet, 20\% on Pointnet++ SSG, 24\% on Pointnet++ MSG, 20\% on DGCNN, and 33\% on CurveNet. The untargeted ASR also boosted by 27\%. We also found 3D-TImay slight improve the perturbation size (Figure \ref{RMSE}). We think this is necessary to resist environmental changes. We also show the influence of hyperparameters in Figure \ref{fig-iteration}.   


\begin{figure}[hbtp]
            	\centering
            		\includegraphics[width=0.4\textwidth]{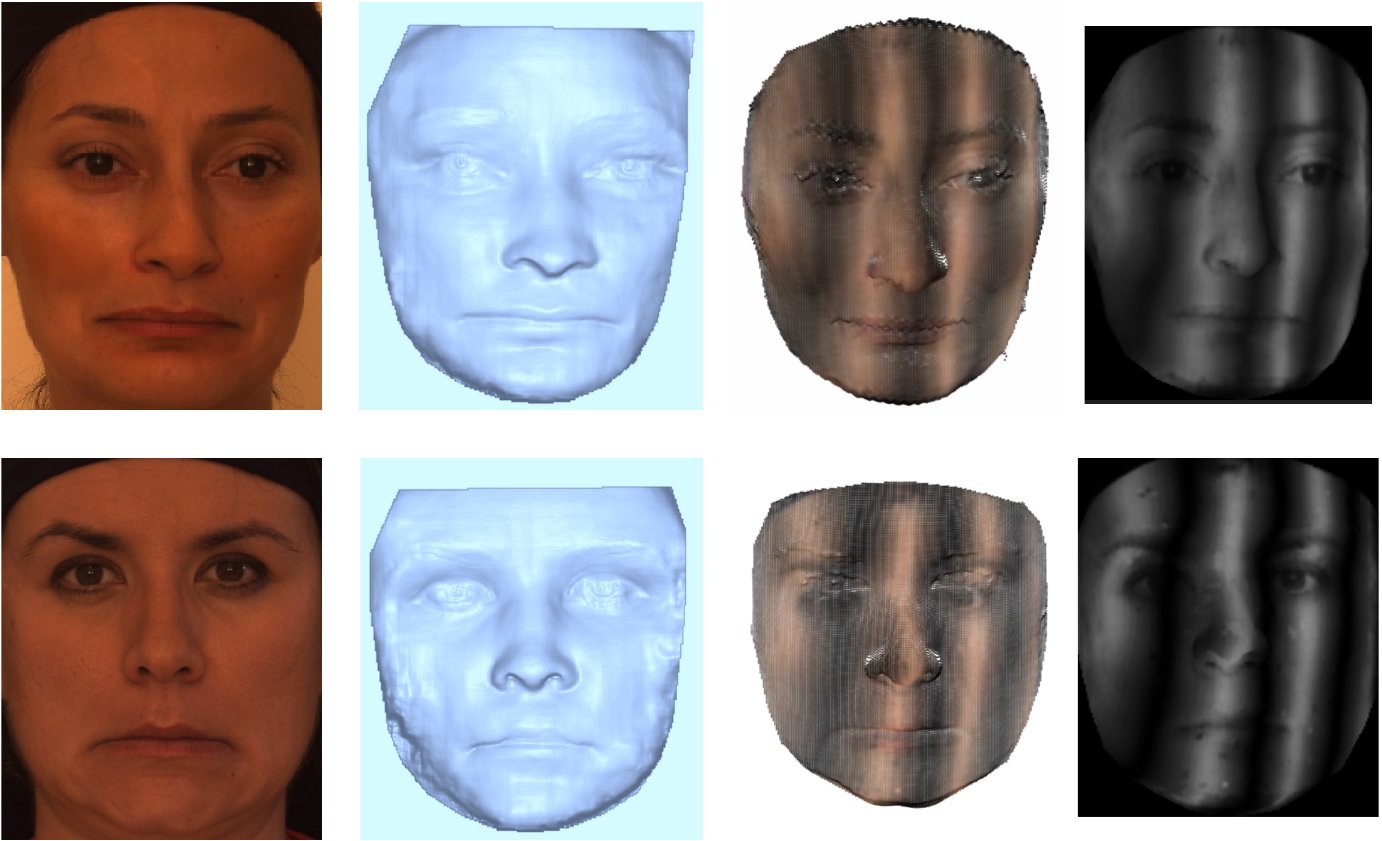}
            		\caption{\textbf{From left to right}: the face images, the depth maps, the 3D faces rendered by the Lambertian model, and modulated fringe images. \textbf{Top row}: normal images. \textbf{Bottom row}: attacked images. }         \label{Lamber}
            	\end{figure}

              \paragraph{Qualitative results of Lambertian model}
                    As revealed in Figure \ref{Lamber}, the Lambertian model can generate photorealistic rendering results for relighting process. These two faces are from the Bosphorus dataset. From left to right are the face images, the ground truth depth maps, the 3D point clouds rendered by the Lambertian model and the 2D images rendered from the 3D models.
              
                   \begin{figure}[h]
            	\centering
            	\begin{subfigure}{0.48\linewidth}
            		\centering
            		\includegraphics[width=1\linewidth]{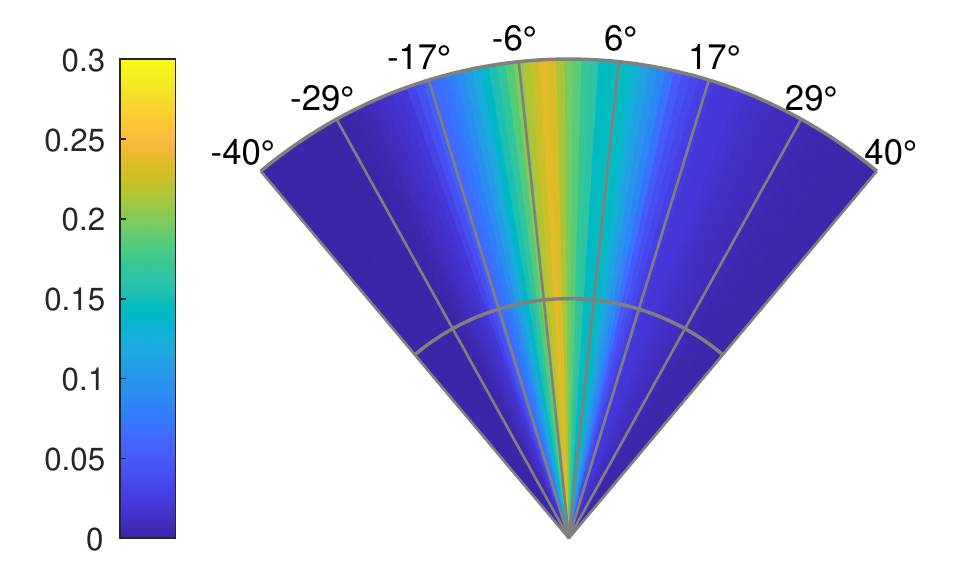}
            		\caption{Without 3D-TI}
            		\label{without3DTI}
            	\end{subfigure}
            	\centering
            	\begin{subfigure}{0.48\linewidth}
            		\centering
            		\includegraphics[width=1\linewidth]{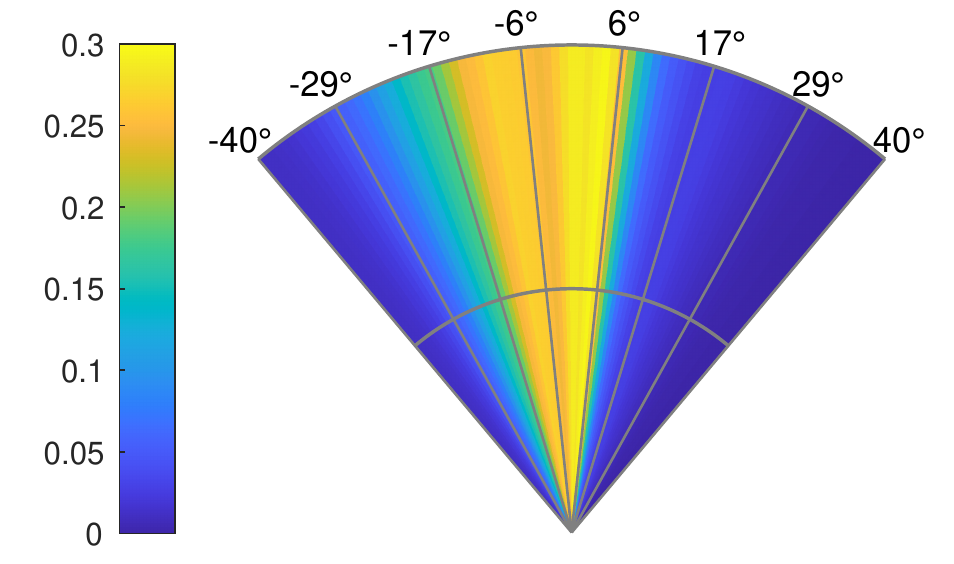}
            		\caption{With 3D-TI}
            		\label{with3DTI}
            	\end{subfigure}

            	\caption{The adversarial examples' robustness against face rotations. We generate perturbations on Pointnet and then add them to the rotated point cloud. We plot the predictions on the target label without (left) and with (right) 3D-TI module.}
            	\label{Fig-hotmap}
            \end{figure}

             \paragraph{Effect of random 3D rotations}
             To evaluate the effectiveness of the 3D transformation-invariant loss (3D-TI), we randomly rotate the original point cloud at different angles and add the same perturbations to the rotated point clouds. Then we compare the predictions on the target label after the softmax layer. As shown in Fig.\ref{Fig-hotmap}, with the 3D-TI module, the prediction is more robust to rotations.


             \begin{figure}[htbp]
            	\centering
            	\begin{subfigure}{0.48\linewidth}
            		\centering
            		\includegraphics[width=1\linewidth]{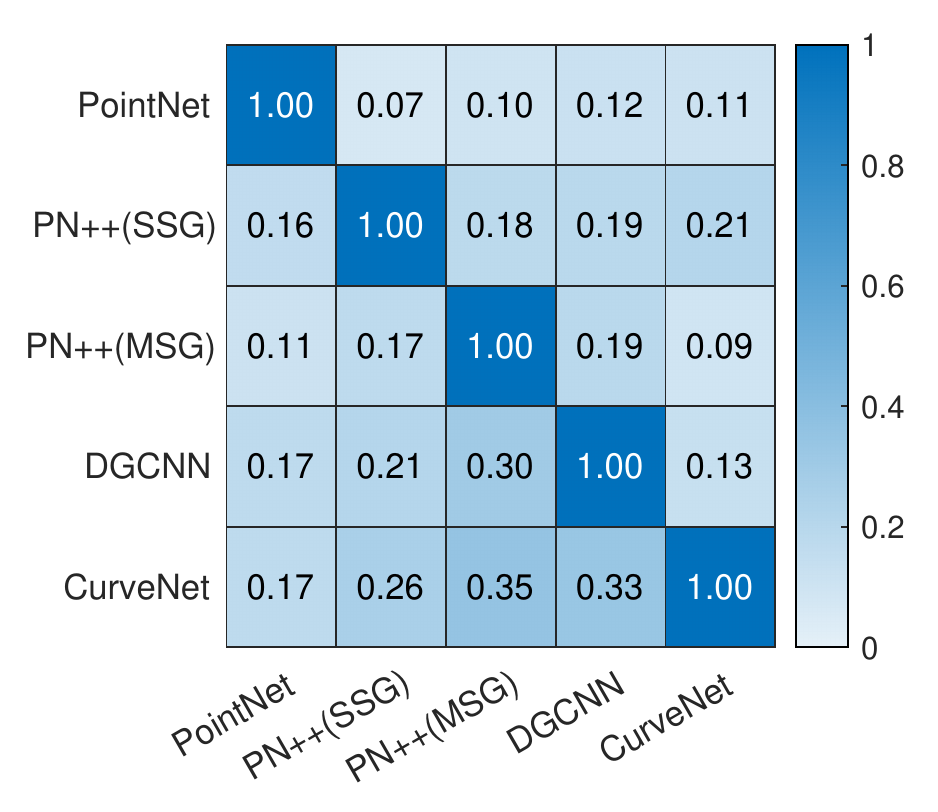}
            		\caption{C\&W}
            		\label{transfer1}
            	\end{subfigure}
            	\centering
            	\begin{subfigure}{0.48\linewidth}
            		\centering
            		\includegraphics[width=1\linewidth]{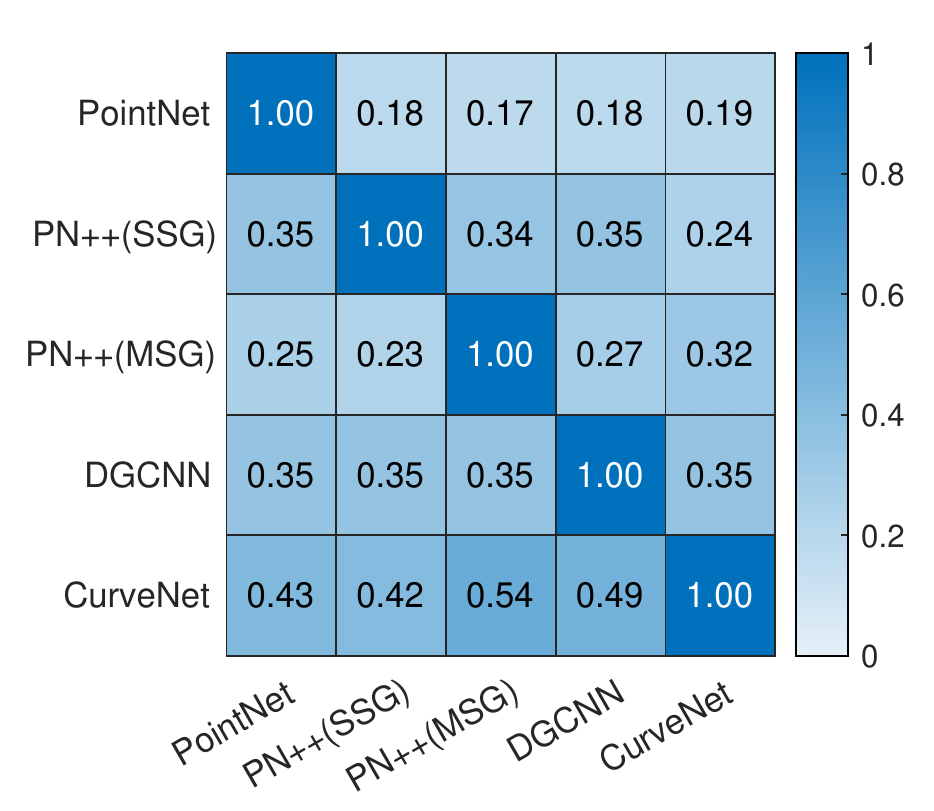}
            		\caption{Ours}
            		\label{tranfer2}
            	\end{subfigure}
             \caption{The transferability of dodging attacks of C\&W and ours. The vertical axis is the shadow model. The horizontal axis is the victim model.}
             \label{Fig-transfer}    
             \end{figure}
            
            \paragraph{Transferability of adversarial examples}
                Previous work has shown that random 2D transformations can improve the transferability of adversarial images \cite{xie2019improving}. We find that 3D-TI loss can also improve the transferability of 3D adversarial examples, especially for dodging attacks. We show the transferability of dodging attacks of C\&W and ours in Table \ref{Fig-transfer}. Moreover, we find that transferability on face datasets is higher than objects dataset like ModelNet40 \cite{wu20153d}. We think this is because the latent features of different faces are more similar to each other compared with different objects.
 
\section{Conclusion}
    \label{conclusion}
    This paper proposes physical-realizable end-to-end adversarial attacks against 3D face recognition. To involve the complex face reflection process in attack pipelines, we model the projection and reflection through the Lambertian model. We also utilize 3D-TI loss and sensitivity maps to improve the attack robustness and imperceptibility. We evaluate our attack on state-of-the-art 3D deep learning models. Quantitative and qualitative analysis shows that our attack can successfully attack the real-world system with very few perturbations. Our end-to-end attacks are not only useful for structured light cameras but also shed light on future real-world attacks that involve face relighting.

{\small

\bibliographystyle{ieee_fullname}

\begin{thebibliography}{10}\itemsep=-1pt

\bibitem{basri2003lambertian}
Ronen Basri and David~W Jacobs.
\newblock Lambertian reflectance and linear subspaces.
\newblock {\em IEEE transactions on pattern analysis and machine intelligence},
  25(2):218--233, 2003.

\bibitem{cao2019adversarial}
Yulong Cao, Chaowei Xiao, Benjamin Cyr, Yimeng Zhou, Won Park, Sara Rampazzi,
  Qi~Alfred Chen, Kevin Fu, and Z~Morley Mao.
\newblock Adversarial sensor attack on lidar-based perception in autonomous
  driving.
\newblock In {\em Proceedings of the ACM SIGSAC conference on computer and
  communications security}, pages 2267--2281, 2019.

\bibitem{carlini2017towards}
Nicholas Carlini and David Wagner.
\newblock Towards evaluating the robustness of neural networks.
\newblock In {\em 2017 ieee symposium on security and privacy (sp)}, pages
  39--57. Ieee, 2017.

\bibitem{dong2019efficient}
Yinpeng Dong, Hang Su, Baoyuan Wu, Zhifeng Li, Wei Liu, Tong Zhang, and Jun
  Zhu.
\newblock Efficient decision-based black-box adversarial attacks on face
  recognition.
\newblock In {\em Proceedings of the IEEE/CVF Conference on Computer Vision and
  Pattern Recognition}, pages 7714--7722, 2019.

\bibitem{951529}
H. Farid.
\newblock Blind inverse gamma correction.
\newblock {\em IEEE Transactions on Image Processing}, 10(10):1428--1433, 2001.

\bibitem{feng2019fringe}
Shijie Feng, Qian Chen, Guohua Gu, Tianyang Tao, Liang Zhang, Yan Hu, Wei Yin,
  and Chao Zuo.
\newblock Fringe pattern analysis using deep learning.
\newblock {\em Advanced Photonics}, 1(2):025001, 2019.

\bibitem{feng2019micro}
Shijie Feng, Chao Zuo, Wei Yin, Guohua Gu, and Qian Chen.
\newblock Micro deep learning profilometry for high-speed 3d surface imaging.
\newblock {\em Optics and Lasers in Engineering}, 121:416--427, 2019.

\bibitem{feng2021calibration}
Shijie Feng, Chao Zuo, Liang Zhang, Tianyang Tao, Yan Hu, Wei Yin, Jiaming
  Qian, and Qian Chen.
\newblock Calibration of fringe projection profilometry: A comparative review.
\newblock {\em Optics and Lasers in Engineering}, 143:106622, 2021.

\bibitem{gilani2018learning}
Syed~Zulqarnain Gilani and Ajmal Mian.
\newblock Learning from millions of 3d scans for large-scale 3d face
  recognition.
\newblock In {\em Proceedings of the IEEE Conference on Computer Vision and
  Pattern Recognition}, pages 1896--1905, 2018.

\bibitem{gnanasambandam2021optical}
Abhiram Gnanasambandam, Alex~M Sherman, and Stanley~H Chan.
\newblock Optical adversarial attack.
\newblock In {\em Proceedings of the IEEE/CVF International Conference on
  Computer Vision}, pages 92--101, 2021.

\bibitem{hamdi2020advpc}
Abdullah Hamdi, Sara Rojas, Ali Thabet, and Bernard Ghanem.
\newblock Advpc: Transferable adversarial perturbations on 3d point clouds.
\newblock In {\em European Conference on Computer Vision}, pages 241--257.
  Springer, 2020.

\bibitem{huang2022shape}
Qidong Huang, Xiaoyi Dong, Dongdong Chen, Hang Zhou, Weiming Zhang, and Nenghai
  Yu.
\newblock Shape-invariant 3d adversarial point clouds.
\newblock In {\em Proceedings of the IEEE/CVF Conference on Computer Vision and
  Pattern Recognition}, pages 15335--15344, 2022.

\bibitem{kato2018neural}
Hiroharu Kato, Yoshitaka Ushiku, and Tatsuya Harada.
\newblock Neural 3d mesh renderer.
\newblock In {\em Proceedings of the IEEE conference on computer vision and
  pattern recognition}, pages 3907--3916, 2018.

\bibitem{kim2017deep}
Donghyun Kim, Matthias Hernandez, Jongmoo Choi, and G{\'e}rard Medioni.
\newblock Deep 3d face identification.
\newblock In {\em 2017 IEEE international joint conference on biometrics
  (IJCB)}, pages 133--142. IEEE, 2017.

\bibitem{liu2018disentangling}
Feng Liu, Ronghang Zhu, Dan Zeng, Qijun Zhao, and Xiaoming Liu.
\newblock Disentangling features in 3d face shapes for joint face
  reconstruction and recognition.
\newblock In {\em Proceedings of the IEEE conference on computer vision and
  pattern recognition}, pages 5216--5225, 2018.

\bibitem{min2014kinectfacedb}
Rui Min, Neslihan Kose, and Jean-Luc Dugelay.
\newblock Kinectfacedb: A kinect database for face recognition.
\newblock {\em IEEE Transactions on Systems, Man, and Cybernetics: Systems},
  44(11):1534--1548, 2014.

\bibitem{modas2019sparsefool}
Apostolos Modas, Seyed-Mohsen Moosavi-Dezfooli, and Pascal Frossard.
\newblock Sparsefool: a few pixels make a big difference.
\newblock In {\em Proceedings of the IEEE/CVF conference on computer vision and
  pattern recognition}, pages 9087--9096, 2019.

\bibitem{nguyen2020adversarial}
Dinh-Luan Nguyen, Sunpreet~S Arora, Yuhang Wu, and Hao Yang.
\newblock Adversarial light projection attacks on face recognition systems: A
  feasibility study.
\newblock In {\em Proceedings of the IEEE/CVF Conference on Computer Vision and
  Pattern Recognition Workshops}, pages 814--815, 2020.

\bibitem{nichols2018projecting}
Nicole Nichols and Robert Jasper.
\newblock Projecting trouble: Light based adversarial attacks on deep learning
  classifiers.
\newblock {\em arXiv preprint arXiv:1810.10337}, 2018.

\bibitem{Piccirilli2016}
Marco Piccirilli, Gianfranco Doretto, Arun Ross, and Donald Adjeroh.
\newblock A mobile structured light system for 3d face acquisition.
\newblock {\em IEEE Sensors Journal}, 16(7):1854--1855, 2016.

\bibitem{qi2017pointnet}
Charles~R Qi, Hao Su, Kaichun Mo, and Leonidas~J Guibas.
\newblock Pointnet: Deep learning on point sets for 3d classification and
  segmentation.
\newblock In {\em Proceedings of the IEEE conference on computer vision and
  pattern recognition}, pages 652--660, 2017.

\bibitem{qi2017pointnet++}
Charles~Ruizhongtai Qi, Li Yi, Hao Su, and Leonidas~J Guibas.
\newblock Pointnet++: Deep hierarchical feature learning on point sets in a
  metric space.
\newblock {\em Advances in neural information processing systems}, 30, 2017.

\bibitem{qiao2020single}
Gang Qiao, Yiyang Huang, Yiping Song, Huimin Yue, and Yong Liu.
\newblock A single-shot phase retrieval method for phase measuring
  deflectometry based on deep learning.
\newblock {\em Optics Communications}, 476:126303, 2020.

\bibitem{savran2008bosphorus}
Arman Savran, Ne{\c{s}}e Aly{\"u}z, Hamdi Dibeklio{\u{g}}lu, Oya
  {\c{C}}eliktutan, Berk G{\"o}kberk, B{\"u}lent Sankur, and Lale Akarun.
\newblock Bosphorus database for 3d face analysis.
\newblock In {\em European workshop on biometrics and identity management},
  pages 47--56. Springer, 2008.

\bibitem{sengupta2018sfsnet}
Soumyadip Sengupta, Angjoo Kanazawa, Carlos~D Castillo, and David~W Jacobs.
\newblock Sfsnet: Learning shape, reflectance and illuminance of facesin the
  wild'.
\newblock In {\em Proceedings of the IEEE conference on computer vision and
  pattern recognition}, pages 6296--6305, 2018.

\bibitem{sharif2016accessorize}
Mahmood Sharif, Sruti Bhagavatula, Lujo Bauer, and Michael~K Reiter.
\newblock Accessorize to a crime: Real and stealthy attacks on state-of-the-art
  face recognition.
\newblock In {\em Proceedings of the 2016 acm sigsac conference on computer and
  communications security}, pages 1528--1540, 2016.

\bibitem{smith2010estimating}
William~AP Smith and Edwin~R Hancock.
\newblock Estimating facial reflectance properties using shape-from-shading.
\newblock {\em International journal of computer vision}, 86(2):152--170, 2010.

\bibitem{tan2019face}
Yang Tan, Hongxin Lin, Zelin Xiao, Shengyong Ding, and Hongyang Chao.
\newblock Face recognition from sequential sparse 3d data via deep
  registration.
\newblock In {\em 2019 International Conference on Biometrics (ICB)}, pages
  1--8. IEEE, 2019.

\bibitem{tsai2020robust}
Tzungyu Tsai, Kaichen Yang, Tsung-Yi Ho, and Yier Jin.
\newblock Robust adversarial objects against deep learning models.
\newblock In {\em Proceedings of the AAAI Conference on Artificial
  Intelligence}, volume~34, pages 954--962, 2020.

\bibitem{tu2020physically}
James Tu, Mengye Ren, Sivabalan Manivasagam, Ming Liang, Bin Yang, Richard Du,
  Frank Cheng, and Raquel Urtasun.
\newblock Physically realizable adversarial examples for lidar object
  detection.
\newblock In {\em Proceedings of the IEEE/CVF Conference on Computer Vision and
  Pattern Recognition}, pages 13716--13725, 2020.

\bibitem{wang2019dynamic}
Yue Wang, Yongbin Sun, Ziwei Liu, Sanjay~E Sarma, Michael~M Bronstein, and
  Justin~M Solomon.
\newblock Dynamic graph cnn for learning on point clouds.
\newblock {\em Acm Transactions On Graphics (tog)}, 38(5):1--12, 2019.

\bibitem{wen2022geometry}
Yuxin Wen, Jiehong Lin, Ke Chen, C.~L.~Philip Chen, and Kui Jia.
\newblock Geometry-aware generation of adversarial point clouds.
\newblock {\em IEEE Transactions on Pattern Analysis and Machine Intelligence},
  44(6):2984--2999, 2022.

\bibitem{wen2020geometry}
Yuxin Wen, Jiehong Lin, Ke Chen, CL~Philip Chen, and Kui Jia.
\newblock Geometry-aware generation of adversarial point clouds.
\newblock {\em IEEE Transactions on Pattern Analysis and Machine Intelligence},
  2020.

\bibitem{Wicker_2019_CVPR}
Matthew Wicker and Marta Kwiatkowska.
\newblock Robustness of 3d deep learning in an adversarial setting.
\newblock In {\em Proceedings of the IEEE/CVF Conference on Computer Vision and
  Pattern Recognition (CVPR)}, June 2019.

\bibitem{worzyk2019physical}
Nils Worzyk, Hendrik Kahlen, and Oliver Kramer.
\newblock Physical adversarial attacks by projecting perturbations.
\newblock In {\em International Conference on Artificial Neural Networks},
  pages 649--659. Springer, 2019.

\bibitem{wu20153d}
Zhirong Wu, Shuran Song, Aditya Khosla, Fisher Yu, Linguang Zhang, Xiaoou Tang,
  and Jianxiong Xiao.
\newblock 3d shapenets: A deep representation for volumetric shapes.
\newblock In {\em Proceedings of the IEEE conference on computer vision and
  pattern recognition}, pages 1912--1920, 2015.

\bibitem{wu2019high}
Zhoujie Wu, Chao Zuo, Wenbo Guo, Tianyang Tao, and Qican Zhang.
\newblock High-speed three-dimensional shape measurement based on cyclic
  complementary gray-code light.
\newblock {\em Optics express}, 27(2):1283--1297, 2019.

\bibitem{xiang2019generating}
Chong Xiang, Charles~R Qi, and Bo Li.
\newblock Generating 3d adversarial point clouds.
\newblock In {\em Proceedings of the IEEE/CVF Conference on Computer Vision and
  Pattern Recognition}, pages 9136--9144, 2019.

\bibitem{xiang2021walk}
Tiange Xiang, Chaoyi Zhang, Yang Song, Jianhui Yu, and Weidong Cai.
\newblock Walk in the cloud: Learning curves for point clouds shape analysis.
\newblock In {\em 2021 IEEE/CVF International Conference on Computer Vision
  (ICCV)}, pages 895--904, 2021.

\bibitem{xiao2021improving}
Zihao Xiao, Xianfeng Gao, Chilin Fu, Yinpeng Dong, Wei Gao, Xiaolu Zhang, Jun
  Zhou, and Jun Zhu.
\newblock Improving transferability of adversarial patches on face recognition
  with generative models.
\newblock In {\em Proceedings of the IEEE/CVF Conference on Computer Vision and
  Pattern Recognition}, pages 11845--11854, 2021.

\bibitem{xie2019improving}
Cihang Xie, Zhishuai Zhang, Yuyin Zhou, Song Bai, Jianyu Wang, Zhou Ren, and
  Alan~L Yuille.
\newblock Improving transferability of adversarial examples with input
  diversity.
\newblock In {\em Proceedings of the IEEE/CVF Conference on Computer Vision and
  Pattern Recognition}, pages 2730--2739, 2019.

\bibitem{9028192}
Yuping Ye, Zhan Song, Junguang Guo, and Yu Qiao.
\newblock Siat-3dfe: A high-resolution 3d facial expression dataset.
\newblock {\em IEEE Access}, 8:48205--48211, 2020.

\bibitem{zheng2019pointcloud}
Tianhang Zheng, Changyou Chen, Junsong Yuan, Bo Li, and Kui Ren.
\newblock Pointcloud saliency maps.
\newblock In {\em Proceedings of the IEEE/CVF International Conference on
  Computer Vision}, pages 1598--1606, 2019.

\bibitem{zhou20183d}
Song Zhou and Sheng Xiao.
\newblock 3d face recognition: a survey.
\newblock {\em Human-centric Computing and Information Sciences}, 8(1):1--27,
  2018.

\bibitem{zhou2007appearance}
Shaohua~Kevin Zhou, Gaurav Aggarwal, Rama Chellappa, and David~W Jacobs.
\newblock Appearance characterization of linear lambertian objects, generalized
  photometric stereo, and illumination-invariant face recognition.
\newblock {\em IEEE Transactions on Pattern Analysis and Machine Intelligence},
  29(2):230--245, 2007.

\bibitem{zhou2018invisible}
Zhe Zhou, Di Tang, Xiaofeng Wang, Weili Han, Xiangyu Liu, and Kehuan Zhang.
\newblock Invisible mask: Practical attacks on face recognition with infrared.
\newblock {\em arXiv preprint arXiv:1803.04683}, 2018.

\bibitem{zuo2022deep}
Chao Zuo, Jiaming Qian, Shijie Feng, Wei Yin, Yixuan Li, Pengfei Fan, Jing Han,
  Kemao Qian, and Qian Chen.
\newblock Deep learning in optical metrology: a review.
\newblock {\em Light: Science \& Applications}, 11(1):1--54, 2022.

\end{thebibliography}
}

\end{document}